\documentclass[letterpaper, 10 pt, conference]{ieeeconf}

\IEEEoverridecommandlockouts  

\usepackage{threeparttable}
\usepackage{tabularx}
\usepackage{multicol}
\usepackage{arydshln}
\usepackage{array}
\usepackage{amsmath}
\usepackage{amssymb}
\usepackage[ruled, vlined, linesnumbered, commentsnumbered]{algorithm2e}
\usepackage{booktabs}
\usepackage{bm}
\usepackage{derivative}
\usepackage{dsfont}
\usepackage{esvect}
\usepackage{etoolbox}
\usepackage{stfloats}
\usepackage{graphicx}
\usepackage{hhline}
\usepackage[bookmarks=true]{hyperref}
\usepackage{multirow}
\usepackage{nccmath}
\usepackage{placeins}
\usepackage{setspace}
\usepackage{siunitx}
\usepackage{tabularx}
\usepackage{upgreek}
\usepackage{verbatim}
\usepackage{wasysym}
\usepackage{xcolor}
\usepackage[normalem]{ulem}
\usepackage[capitalise]{cleveref}
\usepackage{tabularray} 
\usepackage{rotating} 
\usepackage{makecell} 
\usepackage{multirow} 
\usepackage{booktabs} 
\definecolor{WildSand}{rgb}{0.968,0.968,0.968} 
\usepackage{hyperref}
\hypersetup{colorlinks=true,linkcolor=blue,urlcolor=blue}

\appto\TPTnoteSettings{\scriptsize}

\usepackage[letterpaper, left=16.9mm, right=16.9mm, top=20.1mm, bottom=20.1mm]{geometry}

\newcommand{\ie}{\textit{i.e.}}

\newcommand{\name}{{GyroCopter}}
\newcommand{\aoa}{\textit{Rotation-for-bearing}}
\newcommand{\paoa}{\textit{Dual-antenna-bearing}}
\newcommand{\rssi}{\textit{RSSI-only (Ideal)}}
\begin{document}

\title{\LARGE \bf GyroCopter: Differential Bearing Measuring Trajectory Planner for Tracking and Localizing Radio Frequency Sources}

\author{Fei Chen$^{1}$, S. Hamid Rezatofighi$^{2}$  and Damith C. Ranasinghe$^{1}$%
\thanks{This work was supported by the grant LP200301507 from the Australian Research Council (ARC) and the Australian Government’s Research Training Program Scholarship (RTPS).}%
\thanks{$^{1}$Fei Chen and Damith C. Ranasinghe are with the School of Computer \& Mathematical Sciences, The University of Adelaide, SA 5005, Australia.
        {\tt\small fei.chen@adelaide.edu.au, damith.ranasinghe@ adelaide.edu.au}}%
\thanks{$^{2}$S. Hamid Rezatofighi is with the Department of Data Science \& AI at Monash University, VIC 3800, Australia.
        {\tt\small hamid.rezatofighi@ monash.edu}}%
}

\maketitle
\thispagestyle{empty}
\pagestyle{empty}

\begin{abstract}
Autonomous aerial vehicles can provide efficient and effective solutions for radio frequency (RF) source tracking and localizing problems with applications ranging from wildlife conservation to search and rescue operations. Existing lightweight, low-cost, bearing measurements-based methods with a single antenna-receiver sensor system configurations necessitate in situ rotations, leading to substantial measurement acquisition times restricting searchable areas and number of measurements. We propose a \name{} for the task. Our approach plans the trajectory of a multi-rotor unmanned aerial vehicle (UAV) whilst utilizing UAV flight dynamics to execute a constant gyration motion to derive ``pseudo-bearing" measurements to track RF sources. The gyration-based pseudo-bearing approach: i)~significantly reduces the limitations associated with in situ rotation bearing; while ii)~capitalizing on the simplicity, affordability, and lightweight nature of signal strength measurement acquisition hardware to estimate bearings. This method distinguishes itself from other pseudo-bearing approaches by eliminating the need for additional hardware to maintain simplicity, lightweightness and cost-effectiveness. To validate our approach, we derived the optimal rotation speed and conducted extensive simulations and field missions with our \name{} to track and localize multiple RF sources. The results confirm the effectiveness of our method, highlighting its potential as a practical and rapid solution for RF source localization tasks.

\end{abstract}

\IEEEpeerreviewmaketitle

\section{Introduction}
Autonomous radio frequency (RF) source tracking and localization with Unmanned Aerial Vehicles (UAVs) can provide a practical and cost-effective means to service diverse applications such as in conservation biology~\cite{feijofr} and search and rescue~\cite{hong2019,rizk2021}. UAVs ability to navigate unhindered by ground obstacles, coupled with an unobstructed line-of-sight over complex terrains, render them particularly valuable.

Presently, the primary methods employed for RF source tracking with UAVs are: i)~Received Signal Strength Indicator (RSSI) based \cite{korner, hoa2019jofr, vrba2019, pak2023}; and ii)~bearing to radio source based \cite{vander2014, vander2016, UAV_RT, hoffman2023}. The RSSI approach estimates source distance using radio signal strength and benefits from a simple, low-cost receiver system able to accommodate a range of lightweight antennas. This versatility allows for seamless integration of RSSI-based sensor systems into UAVs whilst minimizing added payload and, consequentially, maximizing flight duration and search area~\cite{hoa2019jofr}. However, environmental factors such as terrain variations, as well as the characteristics of radio transmitters and receivers, can substantially affect RSSI measurements. Without accurate modeling of these effects, which is often challenging in practice, trajectory planning to achieve autonomy for tracking and locating RF sources with RSSI methods may yield unreliable results.

\begin{figure}[t!]
    \centering
    \includegraphics[width=0.98\columnwidth]{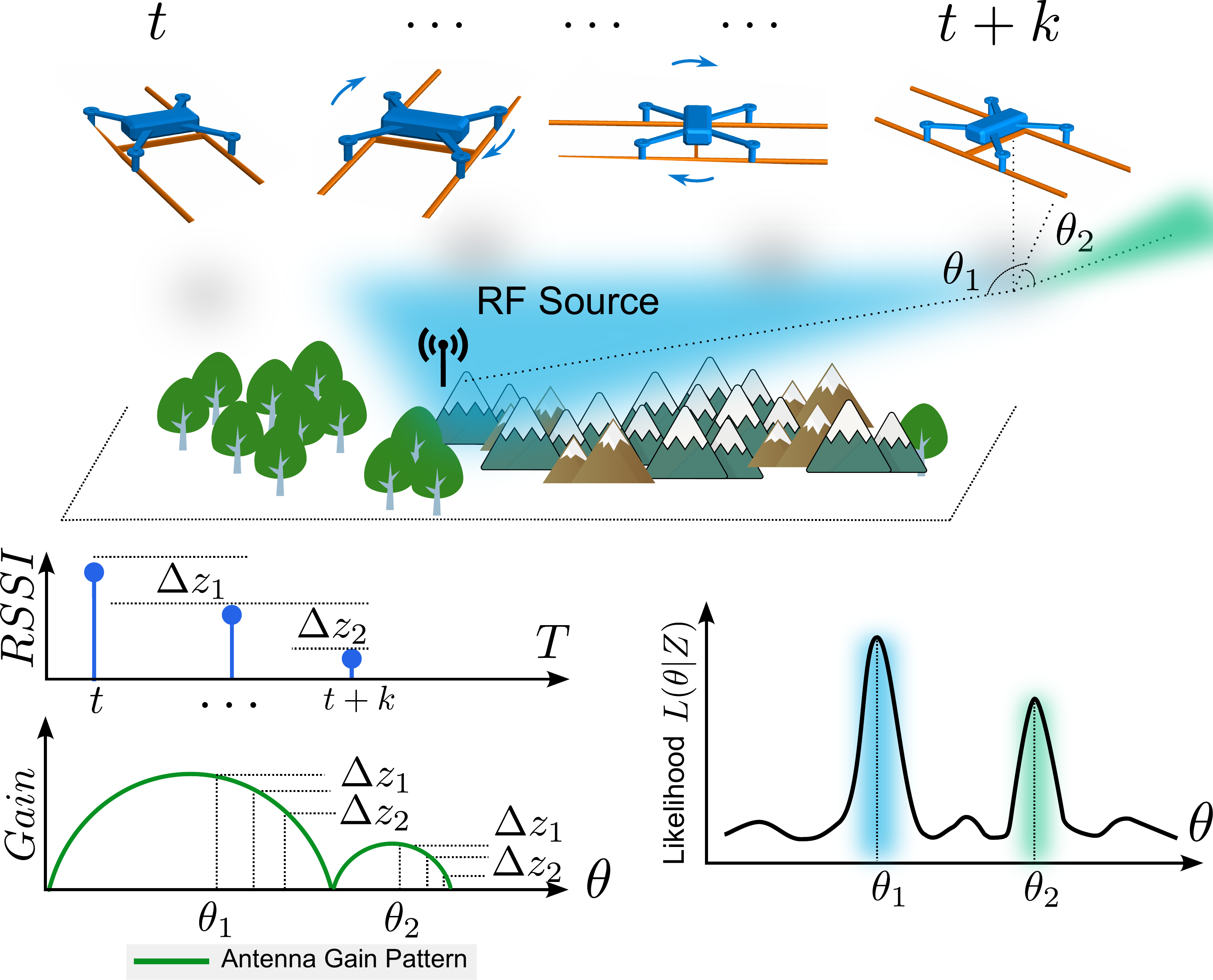}
    \caption{Our proposed Gyrocopter, trajectory planning for tracking and localizing radio frequency (RF) sources. The robotic system integrates \textit{a single directional antenna} undergoing gyrations. This presents different poses of the antenna--from time $t$ to $t+k$---to the RF source to detect changes in received signal strength indicator (RSSI) values to acquire the (\textit{pseudo}-) bearing of an RF source. Importantly, we rely on the differences in a \textit{set} of RSSI measurements, for example, collected from $t$ to $t+k$; therefore, we can remove the impact of environment-dependent parameters on RSSI---often difficult to model/obtain for unknown, complex terrains. See our \name{} demo video is at \textsf{\href{https://youtu.be/OkmmQjD74Us}{https://youtu.be/OkmmQjD74Us}}.}
    \label{fig:uav_figure}
    \vspace{-6mm}
\end{figure}

In contrast, the bearing-only method mitigates the need for knowledge of the RF source transmitter characteristics and signal propagation models. And is more effective in cluttered environments. The most common approach for bearing measurements, due to limited payload capacity, is the ``rotation-for-bearing" method. This involves equipping the UAV with a directional antenna and deducing signal bearings by analyzing the RSSI while the UAV performs a rotational action. 
The signal's bearing can be inferred either by identifying the direction with maximum RSSI \cite{source_seeking_micro_UAV, haluk2023} or by correlating measured RSSI values with the antenna's gain pattern \cite{feijofr,Cliff2018,VonEhr_SDR_AoA2016}. A major drawback of trajectory planning for RF source tracking with this method is the necessity for the UAV to halt and rotate completely to acquire bearing measurements. As a consequence, the UAV needs to undergo frequent accelerations and decelerations, while the number of measurements and search area covered due to the associated time costs become limited.
Further, due to the lengthy measurement acquisition times, approximately $\SI{45}{\second}$. in one study~\cite{Cliff2018}, the method is not suited for localizing mobile objects.

Recent methods incorporating an additional antenna to remove the need to rotate and obtain ``pseudo-bearing" measurements~\cite{two_antenna_pseudo_bearing, hunting_drone}.
However, the approach necessitates multiple antennas and either a multi-port receiver or several receivers, thereby increasing the complexity and weight of the sensor system. Additionally, mounting both a directional and an omnidirectional antenna on a UAV can be challenging at lower frequencies, such as in the VHF (Very High Frequency) band, due to the need for physically larger antennas.

In this paper, we proposed a fully autonomous single-antenna-equipped multi-rotor robot---GyroCopter for RF source tracking and localization tasks. The approach addresses the limitations of the dual-antenna pseudo-bearing method while retaining its advantages in terms of fast measurement acquisitions and robustness of bearing measurements to environmental factors.

Our method leverages the flight dynamics of multi-rotor UAVs, enabling continuous rotation during transit and concurrent RSSI measurements. 
By processing the received RSSI data in brief time frames while the UAV gyrates, we construct pseudo-bearing measurements with a single antenna. Importantly, we demonstrate that the pseudo-bearing measurements thus obtained are also tolerant to variations in RF source transmit power and environmental factors, a desirable attribute of traditional bearing measurements.

In addition, we present a detailed derivation of the optimal rotation speed for the UAV to enhance the accuracy of the pseudo-bearing measurements while maintaining system efficiency.
Unlike the rotation-for-bearing method, our approach significantly reduces the time required for bearing acquisition. Moreover, utilizing just one directional antenna simplifies the receiver system to a single-port receiver. This reduction in complexity not only helps decrease the overall weight of the sensor system and enhance operational efficiency but also allows our proposed method to be seamlessly integrated into existing RSSI-based UAV tracking systems without hardware modifications. Our main contributions are as follows:
\begin{itemize}

    \item We formulate a single antenna-based pseudo-bearing measurement method employing the gyration dynamics of multi-rotor aerial vehicles and derive the optimal rotation speed to maximum localization performance.
    
    \item We integrate our measurement acquisition method with trajectory planning for a tracking algorithm to build a small, fully autonomous robot---\name.
    
    \item We validate the performance of our proposed measurement method and \name{} with extensive simulation and field tests with multiple radio sources.
\end{itemize}

\section{Related Studies}
True bearing measurements based on beam steering require an antenna array and a complex receiver, which are both bulky and expensive to use on small UAVs. The rotation-based strategy provides a practical alternative. Recent studies have sought to reduce measurement times of rotation-based bearing measurements while taking advantage of measurement robustness to variation in transmit power and environmental factors on RSSI. 

The study in~\cite{two_uav_aoa} uses two UAVs equipped with directional antennas. Although the approach does not estimate signal bearing directly, the difference between measured signal strengths is used to infer the position of the source. The method assumes the same signal propagation model for both receivers onboard UAVs.
In~\cite{source_seeking_micro_UAV} and \cite{mono-copter}, a rotating mono-copter uses the direction of maximum signal strength to estimate signal bearing. Notably, a mono-copter is less common, and controlling or trajectory planning for such a vehicle is challenging. In~\cite{two_antenna_pseudo_bearing}, two antennas are configured with a receiver system on a UAV to generate ``pseudo-bearing" measurements to localize an RF source on a fixed flight path. The method supports instantaneous measurements for tracking, but the approach poses practical challenges. The sensor system requires a multichannel radio receiver and installing multiple antennas, which is especially problematic for lower frequency signals due to the larger dimensional and bulky antennas significantly impacting the flight times of small UAVs.

In contrast, we investigate \textit{planning for tracking} with a \textit{continuously gyrating robot} to obtain fast, \textit{pseudo-bearing approximations} using the simplicity of RSSI-based measurement method to reduce the complexity of the receiver, size of the sensor payload, and employ only one antenna whilst benefiting from more robustness attributes of bearing measurements.

\section{GyroCopter Formulation}
Our focus is to provide: i)~a faster yet effective alternative to bearing-based methods to locate radio sources with an aerial robot employing, ii)~a sensor system used in RSSI-based aerial robotic platforms (a single directional antenna and receiver configuration). We provide a detailed formulation of the RF source tracking problem and our approach in the following. 

In summary: we propose using a rolling window of $m$ RSSI measurements gathered from a continuously gyrating robotic platform to obtain near-instantaneous ``pseudo-bearing" measurements. Subsequently, the measurement is used to update the estimated positions (belief state) of RF sources using a recursive Bayesian estimation framework; and we employ the resulting probabilistic belief state to plan the gyrating aerial platform's trajectory to improve tracking accuracy by obtaining more informative measurements.

\subsection{Measurement and Likelihood}
Let $\mathbf{x} = [p^{x}, p^{y}, p^{z}]^{T} \in \mathbb{R}^{3}$ and $\mathbf{u}_{p} = [u^{x}, u^{y}, u^{z}]^{T} \in \mathbb{R}^{3}$ be the location of the radio source and UAV in Cartesian coordinate, respectively; $\mathbf{u} = [\mathbf{u}_{p}; \theta] \in \mathbb{R}^{3}\times [0, 2\pi)$ be the UAV state with its heading angle $\theta$.
Then, the log distance path-loss model for a RSSI measurement can be expressed as~\cite{log_model}:
\begin{equation}
    h(\mathbf{x, u}) = P_{ref}^{(d_{0})} - 10n\log_{10}(\mathbf{d(x, u)}/d_{0}) + G_{r}(\Phi(\mathbf{x, u}))
    \label{eq:rssi_meas}
\end{equation}
where $P_{\text{ref}}^{(d_{0})}$ is the radio source's reference power measured at a distance $d_{0}$, $n$ is the path-loss exponent, $\mathbf{d(x, u)}$ is the Euclidean distance between the radio source and UAV, $G_{r}(\cdot)$ is the received antenna gain, and $\Phi(\mathbf{x, u}) = \operatorname{atan2}(p^{x}-u^{x}, p^{y}-u^{y}) - \theta$ is the azimuth angle \ie{} bearing of the radio source in the UAV's local reference frame.

Typically, the received signal is corrupted by environmental noise.
The effect of can be modeled by Gaussian white noise $\mathbf{w}_{t} \sim \mathcal{N}(0;\sigma^{2})$.
Other phenomena, such as shadowing due to large obstacles such as hills, trees, and buildings, antenna polarization mismatch, or strong multi-path effect, could also contribute to variations in signal attenuation and will be modeled as an unknown and time-varying variable $\bm{\xi}_{t}$.
Therefore, the received signal power at time $t$ can be expressed as environment-dependent terms $h_{d}(\mathbf{x}_{t}, \mathbf{u}_{t})$ and relative bearing-dependent terms $h_{A}(\mathbf{x}_{t}, \mathbf{u}_{t})$ with the addition of noise:
\begin{equation}
    z_{t} = h_{d}(\mathbf{x}_{t}, \mathbf{u}_{t}) + h_{A}(\mathbf{x}_{t}, \mathbf{u}_{t}) + \mathbf{w}_{t}
    \label{eq:full_rssi_meas}
\end{equation}
where $h_{d}(\mathbf{x}_{t}, \mathbf{u}_{t}) = P_{ref}^{(d_{0})} - 10n\log_{10}(\mathbf{d(x, u)}/d_{0}) + \bm{\xi}_{t}$ and $h_{A}(\mathbf{x}_{t}, \mathbf{u}_{t}) = G_{r}(\Phi(\mathbf{x}_{t}, \mathbf{u}_{t}))$

In reality, the exact value of reference power, path-loss exponent, and environmental-dependent loss is usually difficult to determine. This greatly limits the scenario where the received signal strength-based tracking method can be applied.
To address this issue, a normalization step can be carried out to eliminate the environmental-dependent term $h_{d}(\cdot, \cdot)$.

Given $m$ RSSI measurements $\mathbf{z}_{t}=[z_{t-m+1}, \ldots, z_{t}]^{T}$ received at corresponding UAV state $\mathbf{U}_{t}=[\mathbf{u}_{t-m+1}, \ldots, \mathbf{u}_{t}]^{T}$, we consider using the finite differences between each successive RSSI measurement for normalization. Now,
\begin{equation}
    \mathbf{Z}_{t} = [\Delta[z_{t-m+2}], \ldots, \Delta[z_{t}]]^{T}
    \label{eq:diff_norm_eqn}
\end{equation}
where $\Delta[z_{k}] = z_{k} - z_{k-1}$.

In practice, if the measurement sampling period is sufficiently small, it is reasonable to assume that the radio source's transmit power and environmental effects that affect the received signal strength remain constant.
Then, substitute \eqref{eq:full_rssi_meas} into \eqref{eq:diff_norm_eqn}, the measurement function for the normalized measurement can be obtained as:
\begin{equation}
    \mathbf{Z}_{t} = h_{m}(\mathbf{X}_{t}, \mathbf{U}_{t}) + \mathbf{W}_{t}
    \label{eq:norm_meas_func}
\end{equation}

where
\begin{align}
    &h_{m}(\mathbf{X}_{t}, \mathbf{U}_{t}) = \nonumber\\
    &\begin{bmatrix}
    G_{r}(\Phi(\mathbf{x}_{t-m+2}, \mathbf{u}_{t-m+2})) - G_{r}(\Phi(\mathbf{x}_{t-m+1}, \mathbf{u}_{t-m+1})) \\
    \ldots \\
    G_{r}(\Phi(\mathbf{x}_{t}, \mathbf{u}_{t})) - G_{r}(\Phi(\mathbf{x}_{t-1}, \mathbf{u}_{t-1}))\\
    \end{bmatrix}  \label{eq:meas_func} \\     
    &\mathbf{W}_{t} = [\mathbf{w}_{t-m+2} - \mathbf{w}_{t-m+1}, \ldots, \mathbf{w}_{t} - \mathbf{w}_{t-1}]^{T}
\end{align}

The associated likelihood function can then be expressed as:
\begin{equation}
    \mathcal{L}(\mathbf{Z}_{t}|\mathbf{X}, \mathbf{U}) = \mathcal{N}(\mathbf{Z}_{t};h_{m}(\mathbf{X}, \mathbf{U}), \bm{\Sigma}) 
    \label{eq:norm_likelihood}
\end{equation}
\vspace{-1mm}
where:
\begin{align}
\begin{split}
    \bm{\Sigma} = &\mathbb{E}\{\mathbf{W}_{t}^{2}\} = [\Sigma_{ij}], \\
    &\Sigma_{ij}=
    \begin{cases}
        2\sigma^{2} &~\text{if}~i=j \\
        -\sigma^{2} &~\text{if}~i+1=j~\text{or}~j+1=i \\
        0 &~\text{otherwise}
    \end{cases}
\end{split}
\end{align}

\subsection{Object State Estimation}
With measurement model \eqref{eq:norm_meas_func} and likelihood function \eqref{eq:norm_likelihood}, a Bayesian filtering method can be applied to estimate the true state of radio sources.
Due to the non-linearity in the measurement function, we implement a particle filter to estimate radio sources' state (position). 
A particle filter uses a set of weighted particles to represent a probability belief density of an object's state. Given particles representing prior object belief at time $t-1$ as  $p(\mathbf{x}_{t-1}|\mathbf{z}_{1:t-1}) = \sum_{i=1}^{N_{p}}w^{(i)}\delta_{\mathbf{x}^{(i)}_{t-1}}(\mathbf{x})$ and $\phi(\cdot|\cdot)$ denoting the object transition density, the posterior belief density $p(\mathbf{x}_{t}|\mathbf{z}_{1:t})$ can be obtained through the standard Bayesian prediction \eqref{eq:bayesian_predict} and update ~\eqref{eq:bayesian_update} steps:

\begin{equation}
    p(\mathbf{x}_{t}|\mathbf{z}_{1:t-1}) = \int \phi(\mathbf{x}_{t}|\mathbf{x}_{t-1})p(\mathbf{x}_{t-1}|\mathbf{z}_{1:t-1})d\mathbf{x}_{t-1}
    \label{eq:bayesian_predict}
\end{equation}

\begin{equation}
    p(\mathbf{x}_{t}|\mathbf{z}_{1:t}) = \frac{\mathcal{L}(\mathbf{z}_{t}|\mathbf{x}_{t})p(\mathbf{x}_{t}|\mathbf{z}_{1:t-1})}{\int \mathcal{L}(\mathbf{z}_{t}|\mathbf{x}_{t})p(\mathbf{x}_{t}|\mathbf{z}_{1:t-1}) d\mathbf{x_{t}}}
    \label{eq:bayesian_update}
\end{equation}

\subsection{Deriving the Optimal Rotation Speed}\label{sec:crlb-derivation}
According to \eqref{eq:meas_func}, the measurement is directly dependent on the change in relative bearing from UAV to a radio source. Due to this property, we will show that the quality of measurement is highly dependent on the rotation speed of the UAV and subsequently derive the optimal value for a given antenna gain pattern. 

Consider a simple scenario where neither the UAV nor the radio source is mobile. It is easy to notice that if the UAV does not rotate then the measurement function, \eqref{eq:meas_func}, will be zero and be \textit{independent of the radio source's state} $\mathbf{u}$ as the bearing does not change. On the other hand, if the UAV rotates too fast, such that $\Phi(\mathbf{x}_{t'}, \mathbf{u})) = \Phi(\mathbf{x}_{t}, \mathbf{u})) + k\cdot 2\pi, k \in \mathbb{N}$, then the outcome will be the same \ie{} it contains no information on the radio source's state $\mathbf{u}$. Therefore, controlling the rotational speed of the UAV is critical for the \name{} methods, as either, a too fast or too slow rotational speed can lead to poor measurement quality.

Here, we investigate a scenario with a UAV rotating at a constant rate, as sudden changes in the UAV's yaw velocity (rotation) could cause instability and increased power consumption in a practical system.
We use the Cram\'er-Rao lower bound (CRLB) \cite{CRLB_rao, CRLB_cramer} to analyze the fundamental limit on the estimation accuracy at different rotation speeds to determine the optimal value.
CRLB for a linear system in the absence of process noise is computed as the inverse of Fisher information matrix (FIM) \cite{CRLB1978}, 
$\mathbf{J}_{t} = [\mathbf{F}_{t-1}^{-1}]^{T}\mathbf{J}_{t-1}\mathbf{F}_{t-1}^{-1} + \mathbf{H}_{t}^{T}\mathbf{R}_{t}^{-1}\mathbf{H}_{t}$, 
where $\mathbf{H}_{t}$ is the Jacobian of nonlinear measurement function $\mathbf{h}_{t}(\cdot)$, \ie{} $\mathbf{H}_{t}=[\nabla_{\mathbf{x}_{t}}\mathbf{h}^{T}(\cdot)]^{T}$ and $\mathbf{F}_{t}$ is the Jacobian of the state dynamic function, $\mathbf{F}_{t}=[\nabla_{\mathbf{x}_{t}}\mathbf{f}^{T}(\cdot)^{T}]$.

\begin{figure}[hbt!]
    \centering
    \includegraphics[width=0.8\columnwidth]{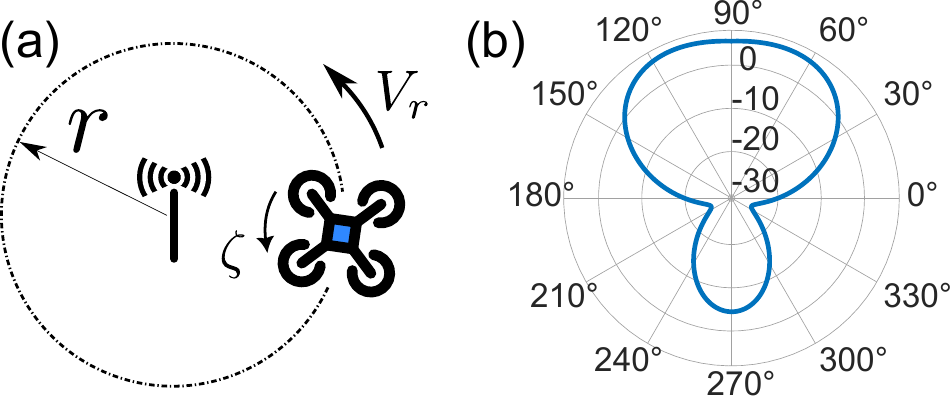}
    \caption{(a) Experiment settings for finding optimal rotation speed; (b) antenna gain pattern of a typical directional H-antenna}
    \label{fig:CRLB_scenario}
\end{figure}

We construct a simplified test scenario that consists of a stationary radio transmitter located at the origin and a UAV equipped with a directional antenna revolving around the radio transmitter at a radius $r$ with constant angular speed $V_{r}$ while self-rotating with a constant angular speed of $\zeta$ degrees per second. Here, measurements are sampled at a period of $\Delta t = \SI{1}{\second}$ by the sensor system onboard the UAV. Fig.~\ref{fig:CRLB_scenario}(a) illustrates the scenario.
Now, let us analyze the optimal rotation speed for the case of $m=2$ measurements and 2D state $\mathbf{x} = [p^{x}, p^{y}]$, $r = \SI{50}{\meter}, V_{r}=\SI{3.6}{\degree/\second}$ and $\sigma=\SI{4}{\decibel}$ where the dynamic model $\mathbf{F} = \mathbf{I}_{2}$, and $\mathbf{I}_{n}$ is a $n\times n$ identity matrix.
Now, the Jacobian $\mathbf{H}_{t}$ is given by:
\begin{equation}
    H_{t} = 
    \begin{bmatrix}
        \pdv*{[G_{r}(\Phi(\mathbf{x}_{t}, \mathbf{u}_{t})) - G_{r}(\Phi(\mathbf{x}_{t-1}, \mathbf{u}_{t-1}))]}{p^{x}_{t}} \\ \pdv*{[G_{r}(\Phi(\mathbf{x}_{t}, \mathbf{u}_{t})) - G_{r}(\Phi(\mathbf{x}_{t-1}, \mathbf{u}_{t-1}))]}{p^{y}_{t}}\\
    \end{bmatrix}^{T} 
    \label{eq:jacobian_h}
\end{equation}

We assume a typical, relatively lightweight 2-element Yagi/H-antenna as in~\cite{nguyen_lavapilot_2020}, and the gain pattern as shown in Fig.~\ref{fig:CRLB_scenario}(b).

\begin{figure}[th!]
    \centering
    \includegraphics[width=1\columnwidth]{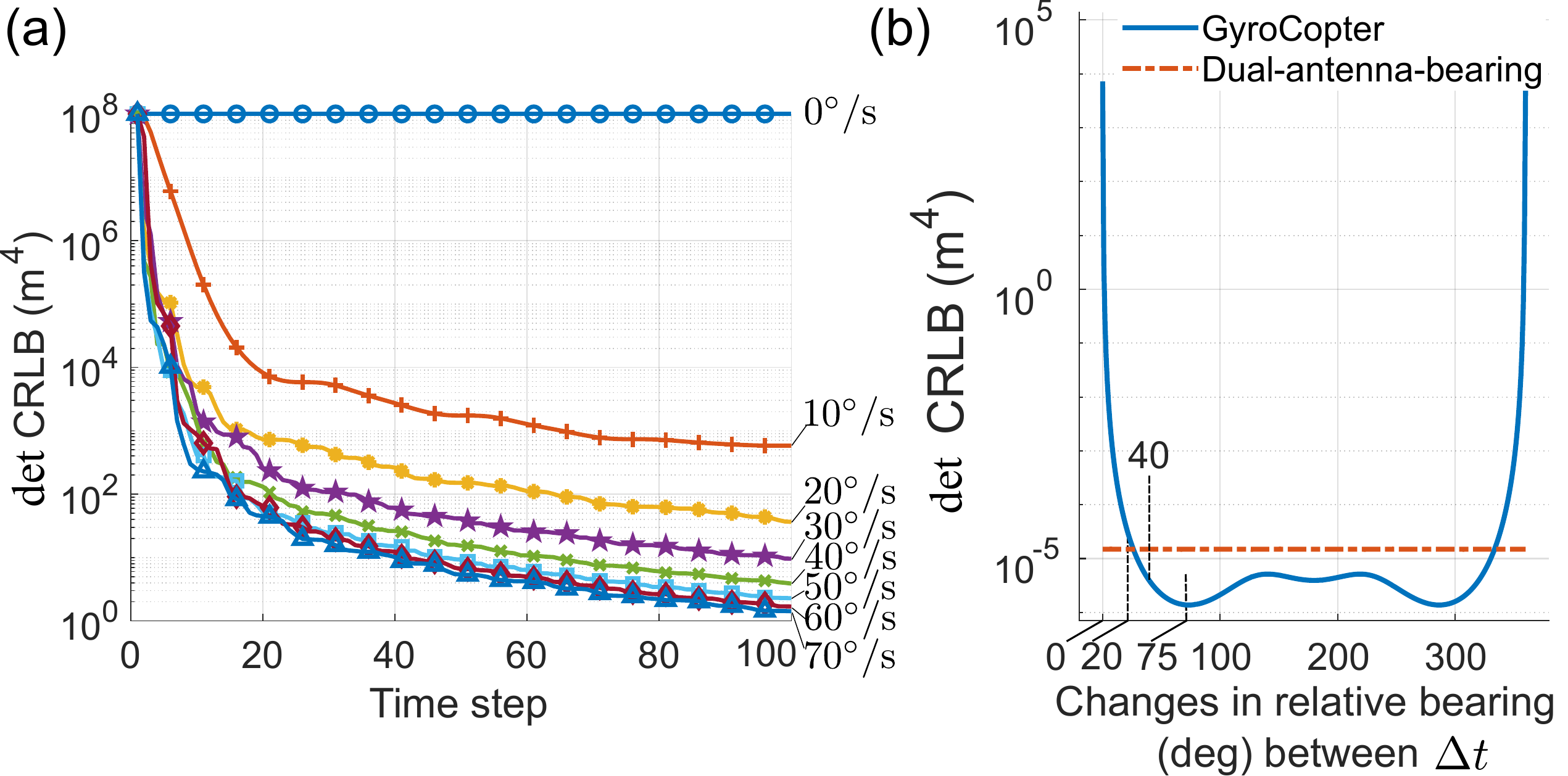}
    \caption{Theoretical analysis of performance. (a)~Determinant of CRLB curves over time with different UAV rotation speed; (b) Determinant of CRLB from the \name{} compared with \paoa{} method's CRLB.}
    \label{fig:CRLB_result}
    \vspace{-4mm}
\end{figure}

Fig.~\ref{fig:CRLB_result}(a) shows the determinant of CRLB values (variance of the estimation error) over time at different rotation speeds.
In all cases except when the UAV does not rotate ($\zeta = \SI{0}{\degree/\second}$), the CRLB value gradually decreases.
This result is intuitive. A UAV that does not rotate will likely receive near identical RSSI measurements, and these measurements are uninformative in the determination of the RF source's bearing.

Fig.~\ref{fig:CRLB_result}(b) shows the steady state of the CRLB determinant values after $100,000$ time steps at different UAV rotation speeds for our proposed \name{} and the \paoa{} methods. Here we use the change in relative bearing over a measurement sampling period $\Delta t$ to implicitly represent the UAV rotation speed to derive a result that is independent of the sampling period. Due to the periodic nature of the rotation motion, the steady state of the CRLB determinant value also has a period of $\ang{360}$ with respect to the rotation angle. For the particular antenna design we investigate with the sampling period $\Delta t$, a minimum CRLB is achieved at approximately $\SI{75}{\degree/\second}$.
In comparison, the \paoa{} method's accuracy is independent of the rotation motion. However, our \name{} is able to achieve better accuracy with a rotation angle between $\Delta t$ greater than $\approx \ang{20}$ as indicated by the lower CRLB value. 
Interestingly, by rotating we are able to increase the difference in the antenna gain over successive measurements, therefore \name's acquisition method is able to obtain more informative measurements. In comparison, the information content from the \paoa{} method is consistently defined by the difference in the gain pattern achieved between the two antennas.

\vspace{3px}
\noindent\textbf{Remark.~}An implication of the results is that increasing the information content from a \paoa{} method requires increasing the directionality of at least one of the antennas. However, this leads to higher weight implication from sensor payloads on-board UAVs. Further, constructing directional antennas, of small form factors, especially at lower frequencies is challenging~\cite{hansen1982}. In comparison, we can achieve such a difference with a single, less directional antenna, by using faster in situ rotations to increase the information content and therefore provide a practical alternative, especially for low frequency RF source tracking.

\subsection{Path Planning}
We propose a computationally efficient task-based planning method.
For RF source localization, a task-based planner provides a computationally lightweight alternative to information-theoretic planning~\cite{nguyen_lavapilot_2020, hoffmann2021}. First, as reported in prior work~\cite{nguyen_lavapilot_2020}, improving localization accuracy benefits from a sensor approaching an RF source to obtain more reliable and higher signal-to-noise (SNR) ratio measurements. Second, planning for the nearest RF source to localize multiple sources rapidly is shown to be an effective strategy to simplify the planning problem~\cite{hoa2019jofr, Cliff2018}. Consequently, we develop a task-based planner with the objective of minimizing the distance between the UAV and the closet estimated RF source location:
\begin{equation}
    \mathbf{a}^{*} = \underset{\mathbf{a}\in\mathcal{A}}{\text{argmin}}~\mathbf{d}(\bar{\mathbf{x}}, f_{uav}(\mathbf{u, a}))
    \label{eq:planner_reward}
\end{equation}

where $\mathbf{a}$ is a control action, $\mathcal{A}$ is an action space, $f_{uav}(\mathbf{u,a})$ is the UAV's state transition function and $\bar{\mathbf{x}} = \int\mathbf{x}\cdot p(\mathbf{x}|\cdot)d\mathbf{x}$ is the expected position of a radio source.
In our planning implementation, we consider two different action spaces; i) discrete; and ii)~continuous. In the continuous space, new velocity vectors are computed to achieve the objective in \eqref{eq:planner_reward}. We describe the two implementations as \textit{Discretized} planner and \textit{Continuous} planner in Algorithm~\ref{alg:planner}.
Here, the nearest source yet to be localized is selected (line 2) before an action is chosen. Under the discrete action space $\mathcal{A}=\{\uparrow, \nearrow,\rightarrow,\ldots\}_{t:t+\Delta T}$, a heading for a $\Delta T$ duration is chosen by the planner to transmit to the flight controller. It has the benefit of reduced computational demands due to infrequent re-planning and the reduced action space.
In contrast, the continuous planner controls UAV via the direct manipulation of its velocity. In this mode a new action is planned every \SI{1}{\second}. 
This enables the UAV to select more optimal trajectories by fully utilizing its maneuverability and thus, potentially reduce the total localization time. Although the computational burden is now increased, we can \textit{still} achieve real-time planning because we considered a \textit{task-based} planning formulation with a simple yet effective and computationally low-cost objective---see \eqref{eq:planner_reward}. Now, the optimal action, $\mathbf{a}^{*}$, found analytically in line 8, maintains a heading to the closest radio source $\bar{\mathbf{x}}^{*}_{t}$.

\begin{algorithm}
\DontPrintSemicolon
\SetKwInOut{Input}{Input}
\SetKwInOut{Output}{Output}
\SetKwInOut{Return}{return}
\Input{$\mathbf{u}_{t}$ UAV state, $v$ UAV speed, $\mathbf{X}=\{\mathbf{x}^{(1)}_{t}\ldots\mathbf{x}^{(n)}_{t}\}$ RF source density set left to localize at time $t$}
\Output{$\mathbf{a}^{*}$ optimal action}

$\mathbf{x}^{*}_{t} = \underset{\mathbf{x}\in\mathbf{X}}{\operatorname{argmin}}~\mathbf{d}(\bar{\mathbf{x}}, \mathbf{u}_{t})$ \tcp*[f]{Closest RF source}\;
    $\mathbf{x}^{*}_{t'} = \text{Predict}(\mathbf{x}^{*}_{t})$ \tcp*[f]{Predict RF source density using Eq.~\eqref{eq:bayesian_predict}}\; 
    \If{\textsc{Discretized}}{
        $\mathcal{A}=\{\uparrow, \nearrow,\rightarrow,\ldots\}_{t:t+\Delta T}$ \;
        $\mathbf{a}^{*}=\underset{\mathbf{a}\in\mathcal{A}}{\text{argmin}}~\mathbf{d}(\bar{\mathbf{x}}^{*}_{t'},  f_{uav}(\mathbf{u}_{t}, \mathbf{a}))$ \tcp*[f]{Eq.~\eqref{eq:planner_reward}}\;
    }
    \ElseIf{\textsc{Continous}}{
        $\mathcal{A}=\{\mathbf{a}: \mathbf{a}\in\mathbb{R}^{2}, ||\mathbf{a}||=v\}$ \tcp*[f]{Velocities} \;
        $\mathbf{a}^{*} = v\cdot(\bar{\mathbf{x}}^{*}_{t^{'}} - \mathbf{u}_{t})/||\bar{\mathbf{x}}^{*}_{t^{'}} - \mathbf{u}_{t}||$ \tcp*[f]{Exact solution for Eq.~\eqref{eq:planner_reward}; vector to $\bar{\mathbf{x}}^{*}_{t}$}\;
    }
\caption{Path Planning (Discrete/Continuous)}\label{alg:planner}
\end{algorithm}

\section{Experiments}
In this section, we describe the simulated and field experiments conducted to validate our proposed system.

\subsection{Simulation Based Comparisons}\label{sec:sim-comp}
We compare our proposed method, in a series of Monte Carlo experiments, with three existing methods: i)~\rssi{}~\cite{hoa2019jofr}; ii)~\paoa{}~\cite{two_antenna_pseudo_bearing}; and iii)~\aoa{}~\cite{Cliff2018}.
The \rssi{} method is used as the baseline method, where the \textit{exact} RSSI measurement model is assumed to be known. In the simulation environment, taking advantage of the ability to control settings in a repeatable manner, we investigate the performance of the methods to track and locate RF sources with increasing mobility. Given the bearing-based methods assume the RF sources to be relatively stationary, this investigation will help appreciate the relative merits (localization accuracy and task completion times) of the approaches as well as predict the operating capabilities of our proposed approach.

\vspace{2mm}
\noindent\textit{Experimental Settings.~}We simulate a scenario with $5$ radio sources placed in a $\SI{1}{\kilo\meter}$ $\times$ $\SI{1}{\kilo\meter}$ search area.
We employed a random walk dynamic model used in~\cite{feijofr,dames2017} with transition density given by:
\begin{equation}
    \phi(\mathbf{x_{k}|x_{k-1}}) = \mathcal{N}(\mathbf{x}_{k};\mathbf{Fx}_{k-1}, \mathbf{Q})
\end{equation}
where $\mathbf{F} = \mathbf{I}_{3}$, $\mathbf{Q} = \sigma_{Q}^{2}\cdot\operatorname{diag}([1, 1, 0]^{T})$, $\sigma_{Q} = 2~\SI{}{\meter}$. The model allows us to generate wandering behaviors for radio sources---typical of radio collared wildlife---to simulate progressively more mobile radio sources by altering the variance, $\sigma_{Q}^{2}$, of the model. Each radio source emits a radio pulse at a frequency of $\SI{1}{\hertz}$ and with $\SI{2}{\decibel}$ of zero-mean Gaussian noise. Particle filter implementations concurrently estimate the state of each RF source. A source is considered localized when the covariance determinant of particle belief on the X-axis and Y-axis is less than $N_{th} = \SI{2e4}{\meter^{4}}$.

To simulate the effect of terrain on the propagation of radio signals, we use the ITU model~\cite[Section 2.2]{ITU-Terrain} to generate the received RF signal at the UAV. The model captures the effect of radio wave diffraction from terrain on all received RSSI measurements. The amount of signal attenuation due to terrain, $P_{d}$, is given by:

    \begin{equation}\label{eq:terrainloss}
        P_{d} = -20\cdot\frac{h}{F_{1}}+10 \text{ (dB)}~~\text{with}~F_{1} = 17.3\cdot\sqrt{\frac{d_{1}d_{2}}{fL}}
    \end{equation}
where $f_{1}$ is the radius of the first Fresnel zone, $L$ is the distance between signal transmitter and receiver; $d_{1}, d_{2}$ are the distances from transmitter and receiver to the blockage in $\SI{}{\kilo\meter}$; and $f$ is the signal frequency in $\SI{}{\giga\hertz}$.

\begin{figure}[htb!]
    \centering
    \includegraphics[width=1.0\columnwidth]{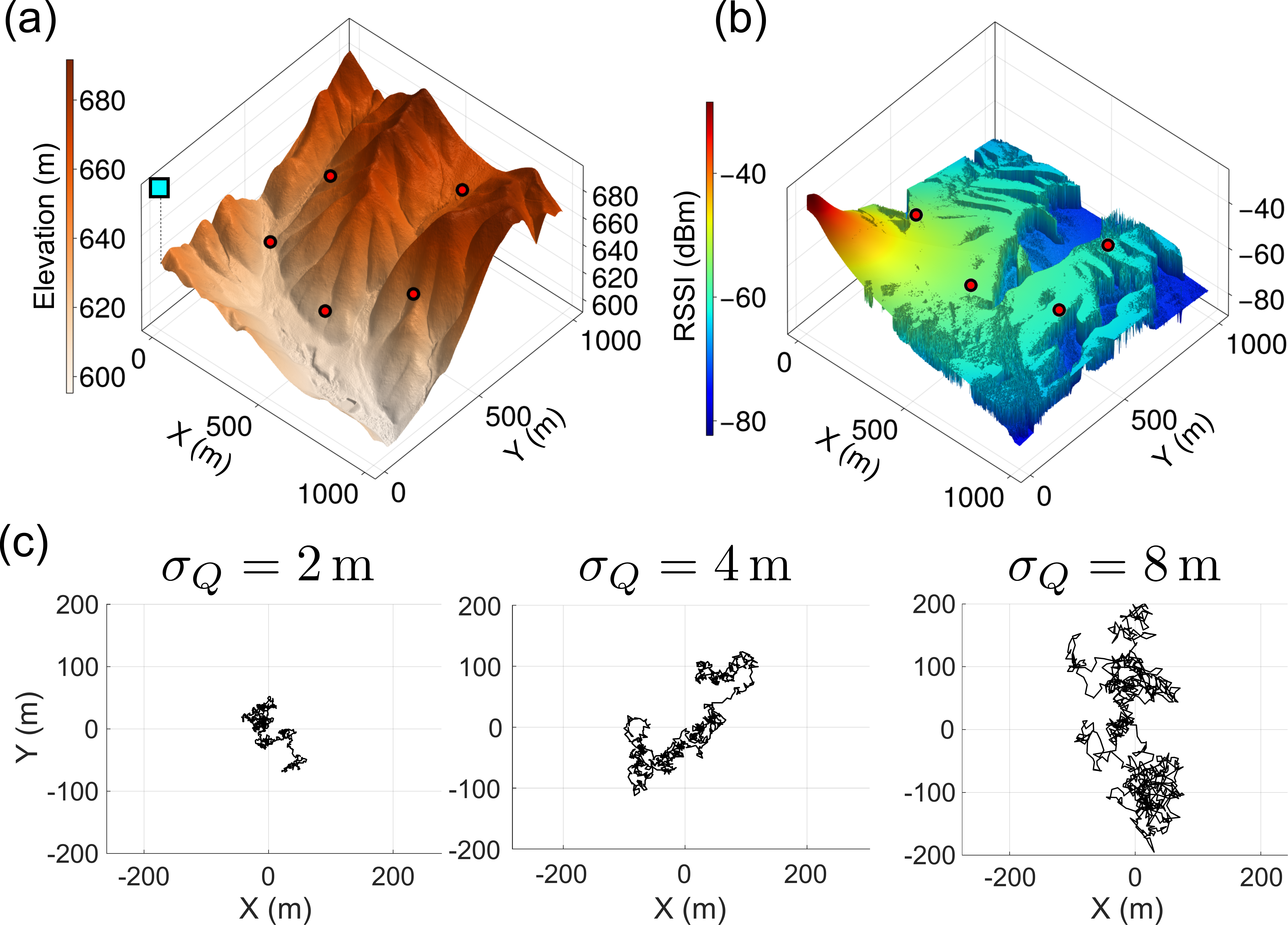}
    \caption{(a) Terrain and RF source placements for the simulation experiments. The initial \name{} and radio sources' locations are marked by $\Box/\ocircle$, respectively. (b) Visualization of the RSSI value generated from the ITU propagation model for a receiver at $\Box$. We can see the impact of the terrain on attenuating the RSSI value. (c) An instance of RF sources' trajectories at different process noise levels, moving from low mobility to high mobility.}
    \label{fig:sim_target_setup}
\end{figure}

In our simulation experiments, for a realistic rendition of terrain variations, we use a digital elevation map of Boorowa, NSW, Australia. 
$P_{truth} = 20~\text{dBm}, n_{truth} = 3$ are used to generate  RSSI measurement, but $P_{model} = 10~\text{dBm}, n_{model} = 2$ are used in the measurement model to \textit{simulate the practical scenario where the characteristics of the signal source are unknown}.
Fig.~\ref{fig:sim_target_setup}(a) illustrates the terrain and RF source placements while Fig.~\ref{fig:sim_target_setup}(b) visualizes signal attenuation impacts we modeled with \eqref{eq:terrainloss} using an RF source moved across the terrain where the receiver is at the initial location ($\Box$).

The simulated UAV travels at a constant $\SI{10}{\meter/\second}$ speed. For our \name{}, we use a constant angular speed of $\SI{40}{\degree/\second}$.
For the bearing only approach (rotation-for-bearing), the UAV takes $\SI{10}{\second}$ to perform a full rotation and obtains a bearing measurement with zero-mean Gaussian noise with a standard deviation of $\ang{9}$. Notably, the rotational speed is significantly faster than the measurement time reported in~\cite{Cliff2018} of $\SI{45}{\second}$ and: i)~aids in generating improved results for the \aoa{} method; as well as ii)~speedup the Monte Carlo-based simulations. More importantly, even whilst using a 4.5$\times$ faster rotation-for-bearing measurement, our proposed approach still demonstrates a significant improvement. A total number of $30$ sets of ground truth paths with randomized initial locations for mobile RF sources were generated, and $35$ Monte Carlo runs for each ground path were conducted to evaluate the performance of different methods.

\begin{table}[htb!]
\resizebox{\columnwidth}{!}{%
\begin{threeparttable} 
    \caption{Simulation result with mobile radio sources.}
    \scriptsize
    \centering
    \begin{tabular}{lll}
    \toprule
    Method & Time (s) & Error (m) \\
    \hline
    \rssi{}~\cite{hoa2019jofr}\tnote{1} & $222\pm48$ & $13.5\pm8.1$ \\
    \hline
    \textbf{Ours} & $\mathbf{297\pm54}$ &  $16.0\pm11.8$\\ 
    \paoa{}~\cite{two_antenna_pseudo_bearing} & $304\pm54$ & $12.8\pm9.0$ \\
    \aoa{}\tnote{2}~\cite{Cliff2018} & $695\pm114$ &  $52.3\pm37.0$\\
    \bottomrule
    \end{tabular}
    \begin{tablenotes}
        \item [1]{Unlike~\cite{hoa2019jofr}, we used an exact measurement model to match the simulated environment to create an \textit{ideal} baseline.}
        \item [2]{With $\SI{10}{\second}$ measurement times, in~\cite{Cliff2018}  rotations needed $\SI{45}{\second}$}.
    \end{tablenotes}
    \label{tab:sim_result_table}
\end{threeparttable}%
}
\end{table}

\begin{figure}[htb!]
    \centering
    \includegraphics[width=\columnwidth]{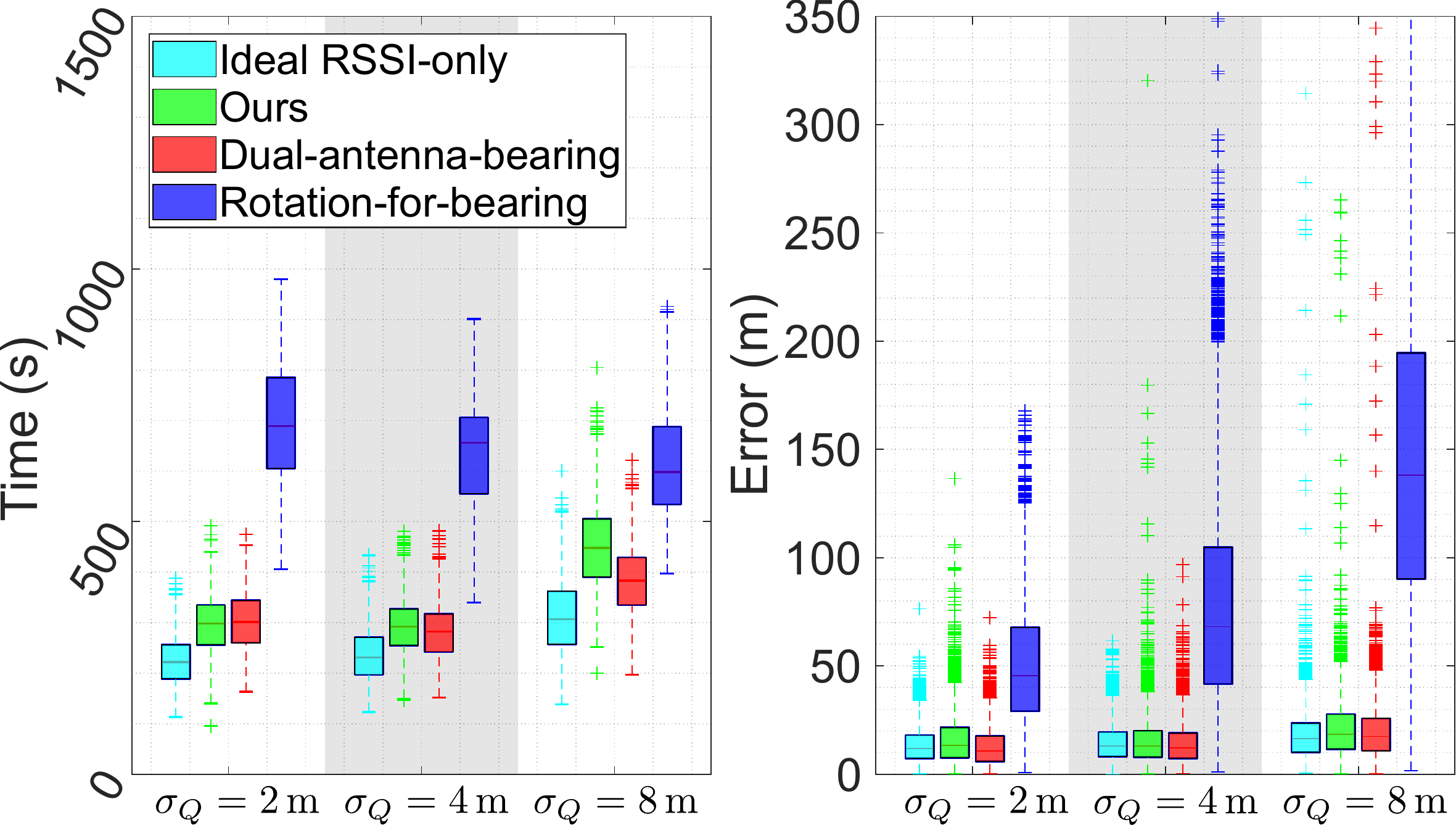}
    \caption{Simulation results comparing the performance of our proposed method for progressing fast-moving radio sources, indicated by higher dynamic model's process noise $\sigma_{Q}$, averaging over $30$ set of randomly generated tracks (with $35$ MC runs each track).}
    \label{fig:var_WD_result}
    \vspace{-4mm}
\end{figure}

\vspace{2mm}
\noindent\textit{Simulation Results.~}
Table~\ref{tab:sim_result_table} shows the mean time and error for the three methods to localize all $5$ radio sources in $100$ Monte-Carlo simulations. 
Our proposed approach significantly outperforms the \aoa{} approach; completing the task in less than $43\%$ of the time and with lower localization error. When compared with the \paoa{} approach, our method is able to complete in approximately the same time duration with similar localization accuracy; the result validates the proposed method's ability to achieve comparable results with simpler hardware requirements satisfied with lighter and smaller---especially at lower frequencies---sensor payload.

Fig.~\ref{fig:var_WD_result} illustrates the total localization time and accuracy achieved by each method with increasingly fast-moving radio sources. Here, the speed of the radio source is directly proportional to its process noise's standard deviation $\sigma_{Q}$. We illustrate sample object trajectories in \textbf{Fig.~\ref{fig:sim_target_setup}(c)}.
As expected, both our proposed \name{} and the  \paoa{} methods are able to significantly outperform the \aoa{} approach in both time and error due to faster measurement acquisition times. 
Interestingly, the faster bearing acquisition methods (\name{} and \paoa{}) require longer duration to complete the localization task with increasing object mobility, albeit still less than the \aoa{} method, on average. 
However, the localization errors of \name{} and \paoa{} remain low across various object mobility settings due to faster, near instantaneous bearing estimates, compared to the \aoa{} method. Notably, the mean localization errors of \name{} remain consistent with the \paoa{} method supported by additional sensor system hardware and are less impacted by the mobility of the RF sources.

\begin{figure*}
    \centering
    \includegraphics[width=0.8\linewidth]{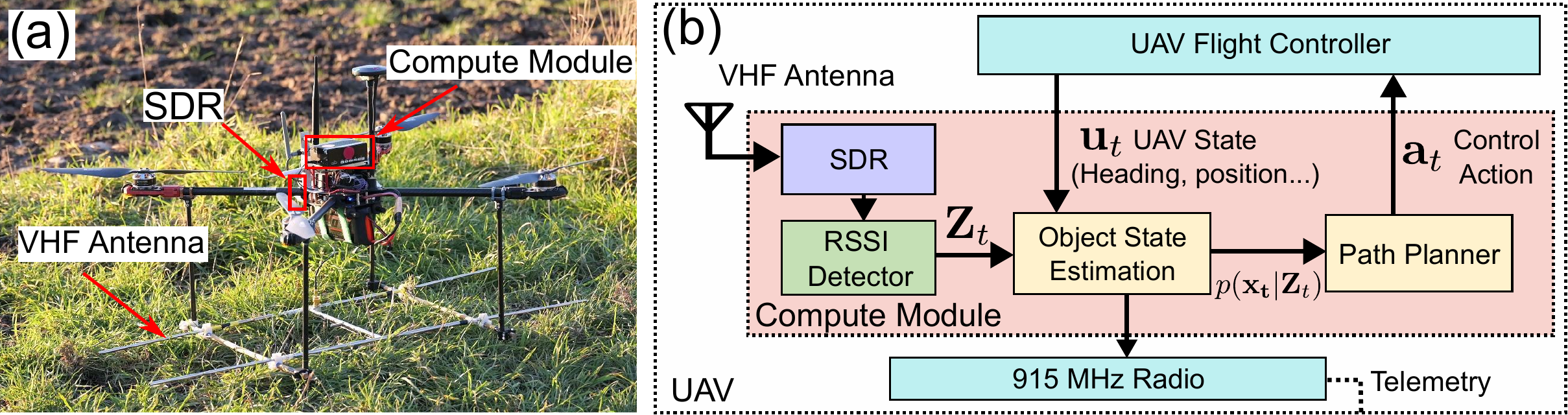}
    \caption{(a) The GyroCopter equipped with a VHF Antenna (H-antenna operating in the VHF range), SDR (USRP B200mini-i) and Compute Module (DJI Manifold 2-C) used in the field experiments. (b) An overview of the system components on the prototype \name. Here the RSSI Detector is implemented using GNUradio building blocks where all of the software components are executed on the Compute Module. See our \name{} demo video is at \textsf{\href{https://youtu.be/OkmmQjD74Us}{https://youtu.be/OkmmQjD74Us}}.}
    \label{fig:system_diagram}
\end{figure*}

\subsection{\name{} Flight Tests}\label{sec:gyro-flight-tests}
We built a sensor system easily deployable in a UAV consisting of a directional antenna, a companion computer, and an SDR (Software Defined Radio).
The UAV used in our system is a custom quad-copter with $\SI{650}{\milli\meter}$ diagonal length frame equipped with a Pixhawk Black flight controller. The sensor system consists of a 2-element H-antenna (Telonics RA-2AK) with a maximum gain of $4$ dBd and $\SI{10}{\decibel}$ front-to-back gain ratio coupled with a USRP B200mini-i SDR receiver. A matched filter-based signal detector is implemented in GNURadio\footnote{An open-source software radio toolkit. \url{https://www.gnuradio.org}} and executed on the companion computer (DJI Manifold 2-C). Localization and planning algorithms are implemented in Julia and are also executed on the companion computer. The total payload weight is approximately $\SI{550}{\gram}$\footnote{We employed a commercial off-the-shelf antenna made of steel for general applications. Consequently, the antenna mass is {\SI{250}{\gram}}. A significant weight reduction can be achieved by using an Aluminum construction.}.
The \name{} and a system overview is shown in Fig.~\ref{fig:system_diagram}.

The radio tags used in this experiment transmit unmodulated On-Off-Keying (OFK) signal with a pulse width of  $\SI{18}{\milli\second}$ and a pulse frequency of $\SI{1}{\hertz}$.

\begin{figure*}[tbh!]
\centering
\begin{minipage}{0.61\columnwidth}
    \centering
    \includegraphics[width=\textwidth]{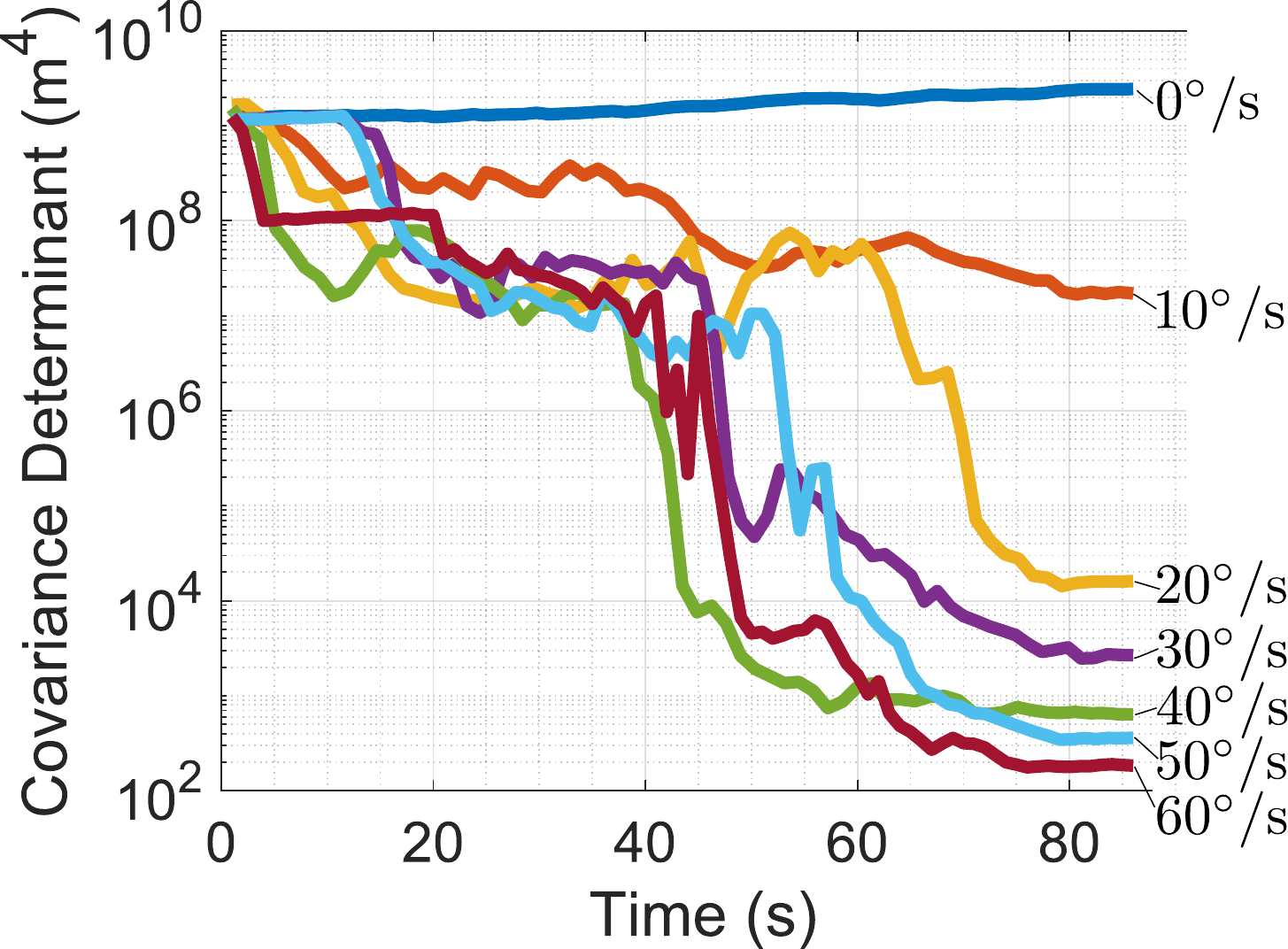}
    \caption{RF source belief density convergence speed testing with angular velocity of $\{0, 10, 20, 30, 40, 50, 60\}^{\circ}/\SI{}{\second}$.}
    \label{fig:rotate_speed_field_result}
\end{minipage}%
\hfill
\begin{minipage}{1.4\columnwidth}
    \centering
    \includegraphics[width=\textwidth]{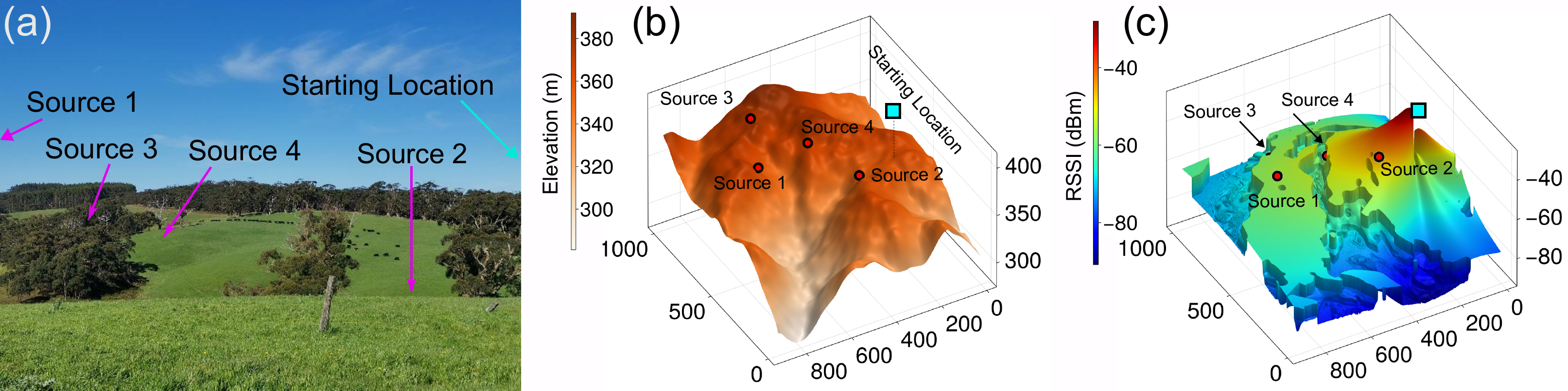}
    \caption{(a) Photograph of field experiment site along with the deployment locations of the radio sources. Notably, radio sources move in and out of line-of-sight from the \name{} during a task. For example, Source~1 and 3 are not in line-of-sight of the UAV at the starting location. (b) Elevation map of the hilly terrain. (c) Receiver coverage. Simulated variations in the received signal strength for a source a various locations for illustrating the significant variations in received signal strength possible from only the terrain variations for a receiver the starting location of the \name{}.
    }
    \label{fig:terrain}
\end{minipage}
\end{figure*}

\noindent\textbf{Rotational Speed and Localization Performance.~}
To establish the impact of the rotation speed and corroborate the results in Section~\ref{sec:sim-comp} illustrated in Fig.~\ref{fig:CRLB_result}, we tasked the \name{} to localize a single radio tag following a pre-defined path at rotational speeds ranging from $\SI{0}{\degree/\second}$ to $\SI{60}{\degree/\second}$ at $\SI{10}{\degree/\second}$ intervals. Fig.~\ref{fig:rotate_speed_field_result} summarizes the field-testing result. As expected, when the rotation speed increases, the (determinant of) covariance of the filter densities reduces faster. The result is consistent with the theoretical result in Fig.~\ref{fig:CRLB_result}(a).

\begin{figure*}[t]
    \centering
    \includegraphics[width=2.03\columnwidth]{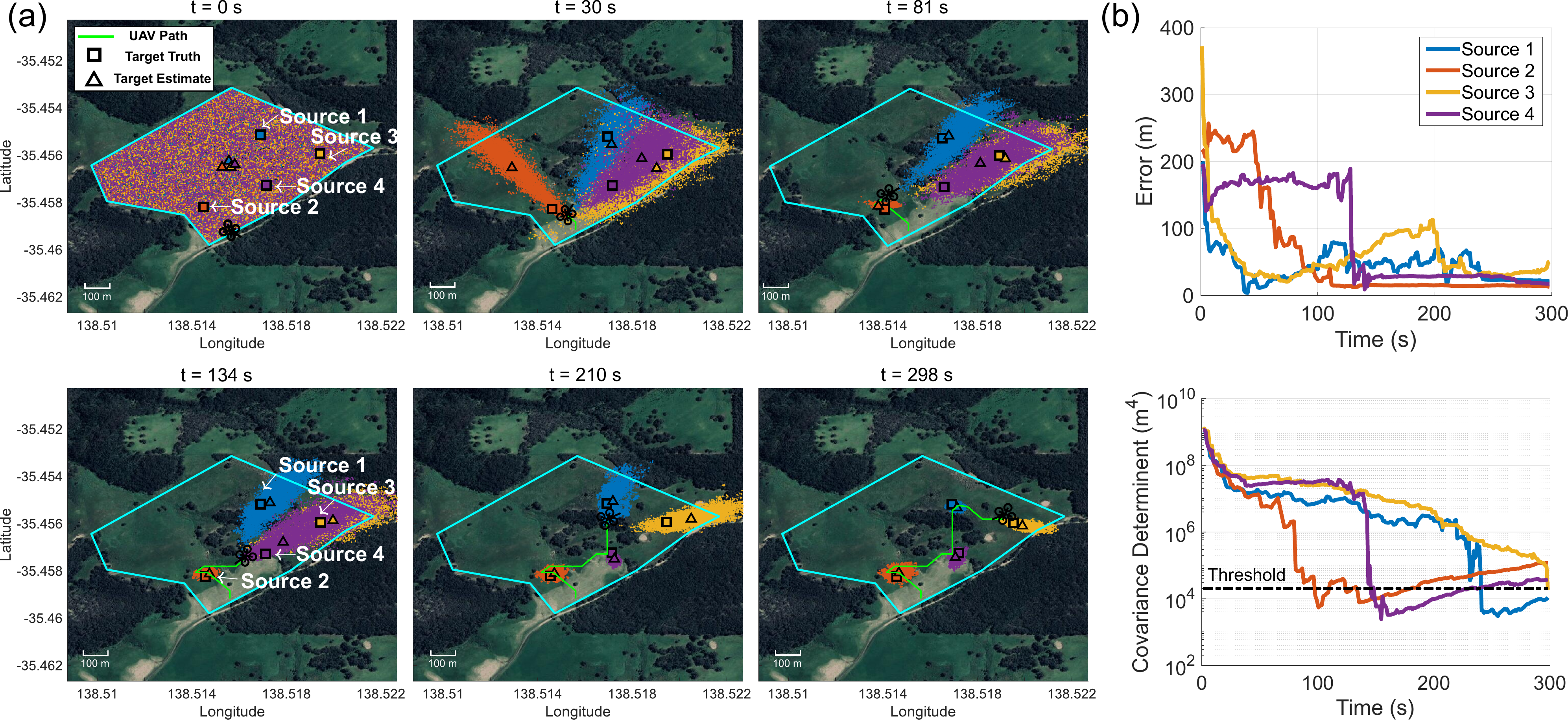}
     \caption{(a) An example field test using the discretized planner for localizing $4$ stationary RF sources. (b) localization error and particle covariance over time. See an example field test with our \name{} at \textsf{\href{https://youtu.be/OkmmQjD74Us}{https://youtu.be/OkmmQjD74Us}}.}
    \label{fig:field_test_instance}
\end{figure*}

\vspace{2px}    
\noindent\textbf{Multiple RF Source Localization Field Tests and Results.~}We conducted flight tests in the Inman Valley, South Australia. The test area was $\SI{40.86}{\hectare}$ in size. Four radio tags were dispersed around the landscape and remained stationary during the full duration of the test to provide a fixed and stable testing setting to help with comparisons. 
The source placement and the elevation map of the test site is shown in Fig.~\ref{fig:terrain}. To compensate for inaccuracy in the antenna model, we used a standard deviation of $\sigma = \SI{5}{\decibel}$ in the measurement model.
The UAV traveled with a speed $v = \SI{5.5}{\meter/\second}$ and rotated at a constant angular speed of $\SI{40}{\degree/\second}$.
Although rotating at the optimal speed, $\SI{75}{\degree/\second}$, determined from our theoretical analysis may improve performance metrics, in practice, we observed very high rotation speeds can impact the UAV's stability. We comfortably operated at $\SI{40}{\degree/\second}$ in our trials, this provided a good balance between performance and UAV stability. Notably, stability can be improved with better alignment of the sensor subsystem's center of gravity with that of the UAV.

We tested our methods with two different types of control actions generated from our planning algorithm Discretized planner and Continuous planner. In both strategies, the planner gyrates the UAV
continuously during the entire mission. With the Discretized planner, we used an action space consisting of $8$ directions with a $\ang{45}$ separation where the UAV travels for $\SI{8}{\second}$ in the chosen heading. A planning step is executed at the end of completing the chosen action. In the Continuous planner, a new velocity vector is computed every $\SI{1}{\second}$.

Fig.\ref{fig:field_test_instance} shows an instance of an experiment where the UAV successfully locates all four radio tags in $\SI{300}{\second}$ by sequentially locating and moving towards each radio tag to improve localization accuracy (shown in the plots depicting localization Error and Covariance). Table~\ref{tab:trial_result} summarizes the field test result for both control methods. A total of $15$ flights were conducted with $11$ using the discrete waypoints (discretized planner) and $5$ using the continuous velocity vector control actions (continuous planner).
All the missions are able to locate the radio sources, consistently. 
While the flights with discretized planner perform reasonably well with good localization accuracy and completion times, the continuous planner is able to outperform it. The reason is because the continuous  action space enables the UAV to select more optimal trajectories by \textit{fully utilizing the maneuverability} of the UAV platform to move to optimal positions to obtain more informative measurements. Although the approach is computationally more intensive, we can observe shorter mission completion times and this can provide the possibility to search larger areas. Notably from Fig.~\ref{fig:terrain}(c), we can observe, even from the starting location, significant variations in attenuation in RF source signals and this can result in missing or not always detecting source signals during a localization task. We capture the mean detection rates from our field experiments in Table~\ref{tab:trial_result}. As expected, due to the differential nature of pseudo-bearing measurement formulation, the \name{} is demonstrated to be robust to such drastic variations in signal attenuation.

\begin{table}[!h]
\centering
\setlength{\extrarowheight}{0pt}
\addtolength{\extrarowheight}{\aboverulesep}
\addtolength{\extrarowheight}{\belowrulesep}
\setlength{\aboverulesep}{0pt}
\setlength{\belowrulesep}{0pt}
\caption{RF source localization field test result with \name{}}
\label{tab:trial_result}
\resizebox{\columnwidth}{!}{%
\begin{tblr}{
  cells = {c},
  column{9} = {WildSand},
  column{2} = {WildSand},
  cell{1}{1} = {r=2}{},
  cell{1}{2} = {r=2}{},
  cell{1}{3} = {c=5}{},
  cell{1}{8} = {r=2}{},
  cell{1}{9} = {r=2}{},
  hline{1,5} = {-}{0.08em},
  hline{2} = {3-7}{},
  hline{3} = {1-8}{0.03em},
  hline{3} = {9}{},
  hline{4} = {-}{},
}
Planner             & \begin{sideways}Trials\end{sideways} & \textbf{Error (m) }$\pm 1\sigma$ &           &           &          &           & \textbf{Time (s)} & {\textbf{Detection}\\\textbf{Rate}} \\
                       &                                      & Source 1                         & Source 2  & Source 3  & Source 4 & Total     &                   &                                     \\
{Discretized} & 10                                   & $35\pm6$                         & $34\pm16$ & $32\pm18$ & $28\pm8$ & $32\pm13$ & $306\pm20$        & 92\%                                \\
{Continuous}  & 5                                    & $33\pm19$                        & $19\pm11$ & $19\pm8$  & $17\pm8$ & $22\pm13$ & $206\pm6$         & 95\%                                
\end{tblr}
}
\end{table}

\vspace{-2mm}
\section{Conclusion} 
\label{sec:conclusion}
In this letter, we propose a single antenna-based pseudo-bearing measurement method employing the gyration dynamics of multi-rotor UAVs to develop an autonomous robot for radio source tracking and localization tasks. Compared to previous approaches, by using only one directional antenna and exploiting the mobility of a multi-rotor UAV, the bearing of the RF signal can be estimated with a simplified radio receiver (reduced payload and costs). The resulting systems are more easily deployed on multi-rotor UAVs. Importantly, compared to rotation-based bearing measuring robotic platforms, ours is significantly faster for tracking and localization tasks. We demonstrate the effectiveness of our proposed \name{} using both simulation-based and field experiments to localize multiple radio sources. In contrast to our proposal to exploit a multicopters' manoeuvrability, future work could consider a servo mechanism to rotate the antenna. Although such an approach increases complexity and payload, it could improve performance by facilitating faster rotations and extend the pseudo-bearing approach to platforms other than multicopters.

\bibliographystyle{IEEEtran}
\bibliography{IEEEabrv,references}
\appendix
\subsection{Alternatives to Exploiting Gyration Dynamics of Multi-Rotor/Multicopter UAVs}

A servo to rotate the antenna is a viable alternative to our proposed approach. It may lead to better UAV stability at faster rotation speeds. However, a number of practical considerations can impact design decisions.

\begin{itemize}
    \item First, rotating the UAV is simple to implement. Current open source autopilot software such as Arducopter/PX4 support standard Mavlink commands, which enable the UAV to rotate while traversing. Thus in situ rotation of a UAV requires no extra software or hardware.
    \item Second, using pseudo-bearing measurements require knowledge of the antenna's heading when an RF pulse is detected. By integrating the antenna to the UAV and using the UAV's heading from its internal instruments along with correction mechanisms, it is straightforward to fetch and use the heading data from the flight controller. Using a servo-based sensor sub-system would need extra equipment for identifying the servo's position, software to smooth and correct such measurements and its integration to the on-board tracking algorithm.
    \item Third, using a servo is actually not simple when the antenna needs to rotate continuously. One difficulty when using a servo is requiring special connectors to route either or both power and RF signals from the rotating platform to the compute module and battery or needing a systems with a built-in rotation mechanism and an antenna. Further, given the desired frequency of operation, an additional challenge is to rotate a structurally large antenna, such as the antenna we employed operating in the VHF frequency band, due to its size, weight and UAV landing gear clearance needed to achieve safe operation.
    \item Fourth, we can expect a determining factor to be if the centre of gravity of the sensor subsystem can remain at the centre of gravity of the UAV during rotations and if the autopilot  needs to make control decisions to stabilize the UAV during rotations executed by a servo mechanism.
\end{itemize}

Importantly, our theoretical analysis in Section~\ref{sec:crlb-derivation} provides a basis to consider possible performance gains. For instance, results in Figure~\ref{fig:CRLB_result}(b) suggest that the overhead of a servo-based sensor subsystem with instrumentation to provide accurate heading information may not yield a significant advantage in our operational settings. 

Consider our trials with radio sources transmitting at $1$~Hz:
\begin{itemize}
    \item The optimal rotation speed is approximately $\SI{75}{\degree/\second}$ according to Figure~\ref{fig:CRLB_result}(b).
    \item The rotation speed we used in our field trials is $\SI{40}{\degree/\second}$.  
\end{itemize}
We perform $100$ Monte Carlo simulations for each rotation speed where the \name{} is tasked to locate $4$ radio sources. 

\begin{table}[!h]
\centering
\caption{RF source localization results at different rotation speeds}
\label{tab:rotate_speed}
\resizebox{\columnwidth}{!}{%
\begin{tabular}{ccc}
\toprule
    Rotation Speed &  Total Time (s) & Error (m) \\
    \midrule
    $\SI{40}{\degree/\second}$ (Setting used in our field trials) &  $248\pm17$ & $10.2\pm6.8$ \\
    $\SI{75}{\degree/\second}$ (Optimal setting from our theoretical analysis) &  $238\pm11$ & $11.3\pm7.5$ \\
    \bottomrule
\end{tabular}%
}
\end{table}

As shown in Table~\ref{tab:rotate_speed}, rotating faster only results in a slight improvement in mean localization time. This is not surprising, since:
\begin{itemize}
    \item The information gain improvement from $\SI{40}{\degree/\second}$ to $\SI{75}{\degree/\second}$ is marginal as seen in Figure~\ref{fig:CRLB_result}(b).
    \item Notably, a rotation speed more than $\SI{20}{\degree/\second}$ is already better than a dual antenna based sensor subsystem approach.
\end{itemize}

Therefore, a system with less moving parts, a lighter payload, less complexity, capable of longer flight times (lower payload, lower computing costs, less power to operate a servo) appear to far outweighs a marginal improvement that may be possible from increasing rotation speed by $20\sim\SI{30}{\degree/\second}$.

Ultimately, using a servo or rotating the UAV are both viable approaches with their own merits and drawbacks. And the best method will likely depend on the specific scenarios, size of the drone, payload capacity of the drone, operating frequency of RF sources, noise in measurements, search range/area, etc. Most importantly, while the choice of implementation can differ, both implementation methods can achieve the goal of generating pseudo-bearing measurements, our reasoning for choosing to rotate the UAV is mainly driven by practical considerations we outlined, theoretical analysis confirming performance improvements over dual antenna-based methods with a simpler sensor subsystem, and the ability to exploit the simple capability of multicopters to perform in situ rotations while moving in different directions without needing extra hardware or software. 

\end{document}